\documentclass{article}

\usepackage{PRIMEarxiv}

\usepackage[utf8]{inputenc} 
\usepackage[T1]{fontenc}    
\usepackage[colorlinks=true,linkcolor=blue,citecolor=blue,urlcolor=blue,filecolor=blue,anchorcolor=blue,breaklinks=true,pdfborder={0 0 0}]{hyperref} 
\usepackage{url}            
\usepackage{booktabs}       
\usepackage{amsfonts}       
\usepackage{nicefrac}       
\usepackage{microtype}      
\usepackage{lipsum}
\usepackage{fancyhdr}       
\usepackage{graphicx}       
\graphicspath{{media/}}     

\usepackage{subcaption}
\usepackage{multirow}
\usepackage{amsmath,amssymb}
\usepackage{amsthm}
\usepackage{mathrsfs}
\usepackage{xcolor}
\usepackage{textcomp}
\usepackage{algorithm}
\usepackage{algorithmicx}
\usepackage{algpseudocode}
\usepackage{listings}
\usepackage{tabularx}
\usepackage{rotating}
\usepackage{makecell}
\usepackage{caption}
\usepackage{float}
\usepackage{placeins}
\usepackage{lineno}

\theoremstyle{plain}

\theoremstyle{definition}

\title{An Integrated Framework of Prompt Engineering and Multidimensional Knowledge Graphs for Legal Dispute Analysis
\thanks{\textit{\underline{Citation}}: 
\textbf{Zhang, M., Zhao, N., Qin, J., Xu, Q., Pan, K., Luo, T. An Integrated Framework of Prompt Engineering and Multidimensional Knowledge Graphs for Legal Dispute Analysis. arXiv:2025.xxxxx}} 
}

\author{
  Mingda Zhang\textsuperscript{1}, Na Zhao\textsuperscript{1,2}, Jianglong Qin\textsuperscript{1,2,*} \\
  School of Software, Yunnan University \\
  Yunnan Provincial Key Laboratory of Software Engineering \\
  Kunming 650500, China \\
  \texttt{\{mingda.zhang, na.zhao, qinjianglong\}@ynu.edu.cn} \\
  \And
  Qing Xu\textsuperscript{3}, Kaiwen Pan\textsuperscript{1,2}, Ting Luo\textsuperscript{1} \\
  \textsuperscript{3}School of Law, Yunnan University \\
  Kunming 650091, China \\
  \texttt{\{qing.xu, kaiwen.pan, ting.luo\}@ynu.edu.cn} \\
}

\begin{document}
\maketitle

\begin{abstract}
Legal dispute analysis is crucial for intelligent legal assistance systems. However, current LLMs face significant challenges in understanding complex legal concepts, maintaining reasoning consistency, and accurately citing legal sources. This research presents a framework combining prompt engineering with multidimensional knowledge graphs to improve LLMs' legal dispute analysis. Specifically, the framework includes a three-stage hierarchical prompt structure (task definition, knowledge background, reasoning guidance) along with a three-layer knowledge graph (legal ontology, representation, instance layers). Additionally, four supporting methods enable precise legal concept retrieval: direct code matching, semantic vector similarity, ontology path reasoning, and lexical segmentation. Through extensive testing, results show major improvements: sensitivity increased by 11.1\%--11.3\%, specificity by 5.4\%--6.0\%, and citation accuracy by 29.5\%--39.7\%. As a result, the framework provides better legal analysis and understanding of judicial logic, thus offering a new technical method for intelligent legal assistance systems.
\end{abstract}


\section{Introduction}
\label{sec1}

Legal dispute analysis, as a complex cognitive task, requires legal professionals to systematically analyze conflicting claims and provide judicial solutions. Lai et al.~\cite{Lai2023a} demonstrated through experimental results that these complex factors cause legal professionals to spend 30-50\% of their time researching applicable laws and precedents. Recent benchmarks for evaluating legal reasoning capabilities have further highlighted these challenges~\cite{Guha2023}. While large language models (LLMs) have shown potential in legal knowledge acquisition and reasoning~\cite{Brown2020}, Wu et al.~\cite{Wu2023} revealed through parameter scale experiments that as models expand from 1 billion to 10 billion parameters, computational resource requirements grow exponentially while performance improvements follow a logarithmic curve, exhibiting diminishing marginal returns. Simultaneously, existing legal language models face key challenges in practical applications, including insufficient depth of legal knowledge representation and limited understanding of specialized legal concepts~\cite{Wang2024}. Various evaluation frameworks have been proposed to assess factual accuracy and reasoning capabilities of these models~\cite{Glanzer2023}.

To address these issues, this research proposes a framework for legal dispute analysis that achieves efficient legal reasoning by integrating prompt engineering with multidimensional knowledge graphs. The framework is based on the concept of selective knowledge node retrieval, which maintains or improves reasoning performance while reducing computational costs~\cite{Lai2023b}. Unlike existing legal AI systems, this framework establishes a comprehensive knowledge-enhanced ecological framework supporting legal decision-making through precise dynamic retrieval of legal concepts, multi-level knowledge representation construction, and professional reasoning path prompt application~\cite{Zhu2024}. Professional enhancement strategies can effectively address capability gaps in large language models regarding legal norm application and case analysis~\cite{Zubaer2023}. Large language models enhanced with feedback from legal professionals can improve legal reasoning ability, but knowledge expression and reasoning path optimization remain challenging~\cite{Sun2023}. From a legal practice regulatory perspective, legal AI systems need to balance innovation with standardization to ensure legal professionalism and ethical standards.

Based on the identified challenges in legal dispute analysis and the design of this framework, this research offers two technical contributions through the integration of prompt engineering and multidimensional knowledge graphs:

\textit{Legal Three-Stage Prompt Engineering Framework:} Breaking the limitations of traditional flat-structure prompt engineering, this framework designs a hierarchical prompt framework consisting of task definition, knowledge background, and reasoning guidance components. Professional legal reasoning path templates have been developed for different types of legal disputes, forming structured chains of thought composed of professional analysis steps~\cite{Wei2022}. Through multi-dimensional evaluation of response quality and dynamic prompt optimization mechanisms, adaptive enhancement is achieved~\cite{Mumford2023}.

\textit{Multidimensional Knowledge Graph and Multidimensional Retrieval Update System:} A three-layer architecture knowledge graph including legal classification ontology, legal representation, and legal instance layers has been constructed, enabling precise dynamic retrieval of legal concepts and knowledge timeliness management~\cite{Hogan2021}. Through four complementary matching strategies and legal concept diversity control mechanisms, the model can consider different legal perspectives~\cite{Li2021}. Additionally, integration with web search technology forms a comprehensive legal knowledge enhancement ecological framework~\cite{Wan2024}.

\section{Related Work}
\label{sec2}

\subsection{Applications and Challenges of Large Language Models in Legal Dispute Analysis}
\label{sec2.1}

Large language models face structural deficiencies in professional legal knowledge representation and reasoning capabilities in the field of legal dispute analysis~\cite{Lai2023b,Brown2020}. Although these models are trained on extensive general text corpora, the high specialization of the legal domain, terminological precision, and jurisdictional differences constitute significant knowledge gaps. Experimental results indicate that while large language models demonstrate certain legal reasoning capabilities even in few-shot learning scenarios, their professional depth remains limited~\cite{Choi2023}. Key challenges facing legal large language models include the high specialization and relative scarcity of legal data, the rapid iteration problem of legal knowledge, and the complexity and multi-layered nature of legal reasoning chains~\cite{Wang2024}. Recent attempts at enhancing legal LLMs have focused on integrating domain-specific knowledge and improving factual grounding capabilities~\cite{Shi2024a}.

To address these challenges, this research constructs an enhanced framework integrating prompt engineering with multidimensional knowledge graphs, achieving efficient legal analysis through precise dynamic retrieval of legal concepts, multi-level knowledge representation construction, and professional reasoning path prompt application, providing novel methodologies for model application in legal dispute analysis~\cite{Guha2023b}. Based on the concept of selective knowledge node retrieval, this framework creates a legal decision knowledge enhancement ecological framework while reducing computational costs and maintaining or improving reasoning performance~\cite{Dhani2021}. This research designs a three-stage hierarchical prompt structure, constructs a three-layer architecture legal knowledge graph, implements precise dynamic retrieval of legal concepts and knowledge timeliness management, and ensures the model considers different legal perspectives~\cite{Jayakumar2023}.

\subsection{Development of Professional Domain Prompt Engineering}
\label{sec2.2}

Prompt engineering, as a key technical method for optimizing large language model outputs, has shown application potential in professional domains but faces applicability limitations in the legal field~\cite{Chen2023,Wei2022}. The "chain of thought" prompting method, by guiding models to explicitly demonstrate the reasoning process, has improved accuracy in complex reasoning tasks, but faces numerous challenges in the legal domain: the highly structured nature of legal reasoning, precision requirements of professional terminology, and standard requirements for legal case citations~\cite{Zhang2023b}. Existing research indicates that while multi-stage prompt frameworks can optimize legal document generation~\cite{Cheng2023}, and structured prompts integrating professional legal knowledge can improve the accuracy of legal judgment prediction~\cite{Zhang2023KL}, there is a lack of systematic prompt optimization methods specifically for legal dispute analysis tasks, particularly integration paths with multidimensional knowledge graphs~\cite{Wang2024a}.

The legal three-stage prompt engineering framework proposed in this research transcends the limitations of traditional prompt engineering's flat structure, designing a hierarchical structure combining task definition, knowledge background, and reasoning guidance to provide comprehensive guidance for legal reasoning~\cite{Trautmann2022}. This research employs a dynamic task identification matching algorithm to precisely map user queries to predefined legal task types; constructs hierarchically structured knowledge backgrounds through legal concept relevance ranking; and provides professional legal reasoning guidance frameworks based on legal reasoning path templates~\cite{Cui2023}. Simultaneously, it designs a dynamic prompt optimization mechanism, including multi-dimensional response quality assessment and adaptive optimization components, enabling dynamic adjustment and quality improvement of prompts, demonstrating advantages in handling complex legal reasoning and multi-dimensional legal analysis.

\subsection{Current Status and Challenges of Multidimensional Knowledge Graphs}
\label{sec2.3}

Multidimensional knowledge graphs, as a core technical foundation for legal knowledge representation and reasoning, face challenges including insufficient knowledge representation granularity, knowledge timeliness issues, and limitations in expressing complex legal relationships~\cite{Liu2023,Li2024}. Technical barriers to legal knowledge graph applications mainly focus on professional requirements for knowledge acquisition and representation, and multi-relational legal knowledge graphs also struggle to fully express complex legal arguments and reasoning chains~\cite{Pan2023}. Recent research indicates that legal knowledge graph construction is transitioning from manual to automated construction~\cite{Xiao2023}, but how to achieve collaborative optimization of knowledge graphs and prompt engineering in legal dispute analysis scenarios remains to be resolved.

The multidimensional knowledge graph designed in this research incorporates both traditional knowledge graph node and relationship representations and multi-granularity concept retrieval mechanisms~\cite{Zheng2021}. This research constructs a three-layer architecture including legal classification ontology, legal representation, and legal instance layers, achieving comprehensive coverage from abstract legal concepts to specific case applications. This architecture conforms to the hierarchical organizational characteristics of legal knowledge and can accurately express hierarchical relationships between legal concepts. Through four complementary matching strategies including legal norm code matching, this research achieves precise dynamic retrieval of legal concepts, establishes a unified retrieval interface covering authoritative legal data sources, ensures the timeliness and reliability of retrieved information, and improves the accuracy and professionalism of legal dispute analysis~\cite{Yang2023}.

\FloatBarrier
\section{Key Technical Design and Implementation of the Legal Dispute Analysis Framework}
\label{sec3}

The legal dispute analysis framework proposed in this research integrates two collaborative core technologies: legal three-stage prompt engineering and multidimensional knowledge graph. Through the integration of these two technologies, a comprehensive legal dispute analysis technical ecological framework is formed, enhancing the entire process from legal concept identification and knowledge acquisition to reasoning guidance. The overall architecture of the framework presents a dual-module collaborative system design concept, where the prompt engineering module is responsible for legal reasoning guidance, the knowledge graph module provides legal knowledge support, and the two interact and integrate at multiple levels.

\begin{figure}[htbp]
\centering
\includegraphics[width=120mm]{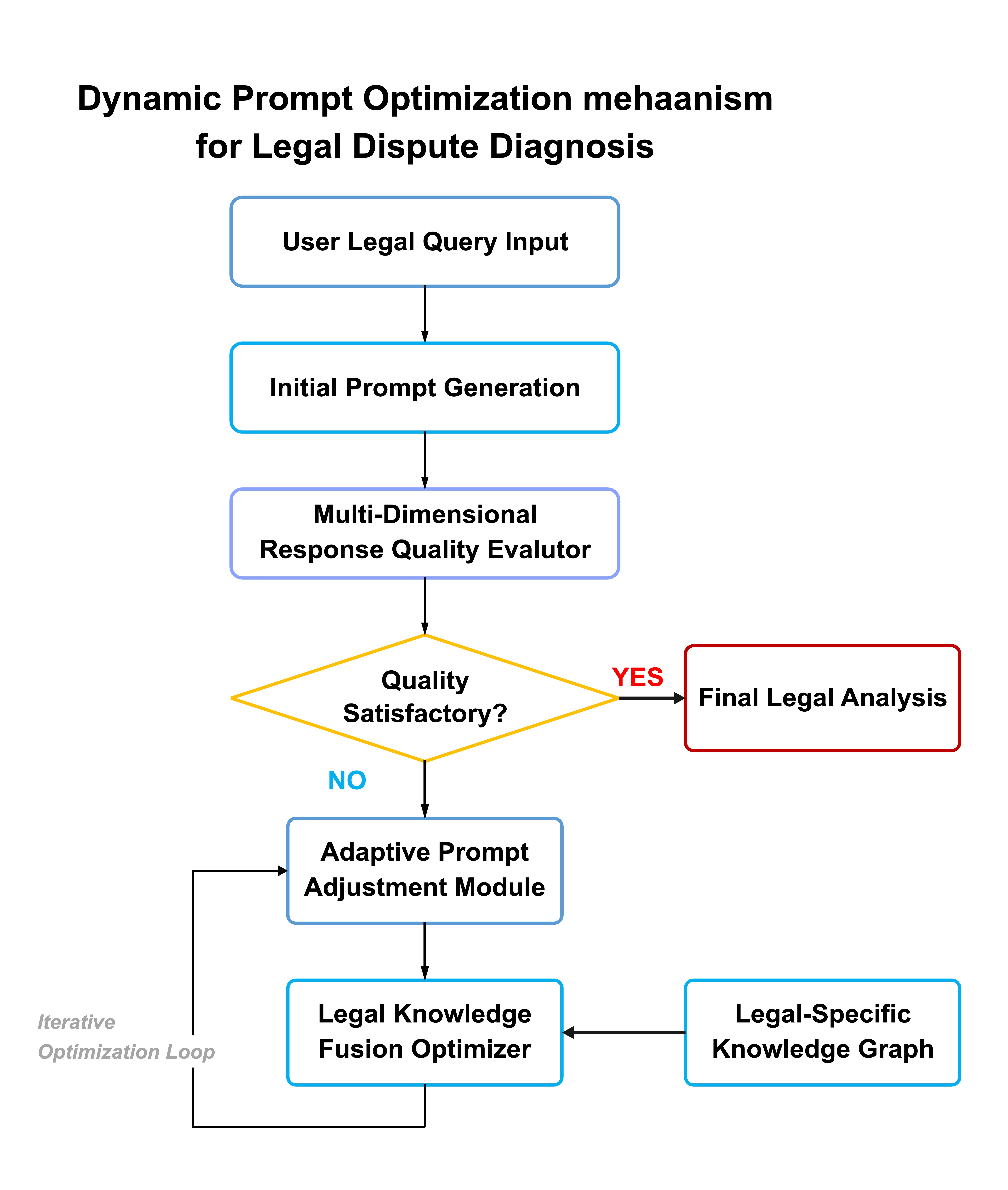}
\caption{Overall Architecture of the Legal Dispute Analysis Framework}
\label{fig1}
\end{figure}

As shown in Figure~\ref{fig1}, the multidimensional knowledge graph on the left and the three-stage prompt engineering on the right constitute the dual core of the system. When a user inputs a legal dispute query, the system first identifies key legal concepts through the multi-granularity concept retrieval component, while activating the knowledge graph query module to obtain relevant legal knowledge. This knowledge is subsequently integrated by the three-stage prompt engineering module, generating structured prompts containing task definition, knowledge background, and reasoning guidance, ultimately guiding the large language model to generate professional legal responses. Figure~\ref{fig1} illustrates the data flow transmission paths between components, reflecting the complete processing flow from user input to system output, as well as the logical connections between functional modules. This dual-core design implements the integration of knowledge retrieval and reasoning guidance, while ensuring the robustness and adaptability of the system when facing complex legal issues.

\subsection{Legal Three-Stage Prompt Engineering Framework}
\label{sec3.1}

Traditional prompt engineering methods generally exhibit problems of single structure, lack of professional legal thinking paths, and insufficient reasoning depth. The three-stage prompt engineering framework proposed in this research transcends the limitations of flat structures, designing a hierarchical structure combining task definition, knowledge background, and reasoning guidance components to effectively guide large language models in professional legal analysis. The effectiveness of structured legal knowledge understanding frameworks has been verified in multiple studies, providing empirical support for the hierarchical prompt structure design of this research.

\subsubsection{Task Definition and Precise Role Positioning Mechanism}
\label{sec3.1.1}

The legal task identification matching algorithm implements precise mapping from queries to professional task templates through multi-dimensional matching mechanisms, with the calculation formula:

\begin{equation}
\begin{split}
M(Q, T_i) = \sum_{j=1}^{n} w_j \cdot \frac{Q_j \cdot T_{ij}}{|Q_j| \cdot |T_{ij}|} \cdot \log\left(\frac{N}{df_j}\right) \cdot (1 + \alpha \cdot SpecTerms(j))
\end{split}
\end{equation}

where $Q$ represents the user's legal query text, $T_i$ is the predefined $i$-th legal task template, $n$ is the number of feature dimensions (including legal domain category, question nature, involved legal provisions, etc.), $Q_j$ and $T_{ij}$ are the representation vectors of the query and template in the $j$-th feature dimension, $w_j$ is the importance weight of the $j$-th feature dimension, determined by legal experts using the Delphi method and satisfying the weight normalization condition, $N$ is the total number of legal documents, $df_j$ is the number of documents containing feature $j$, $SpecTerms(j)$ is the number of professional legal terms contained in feature $j$, and $\alpha$ is an adjustment parameter. This improved weighting scheme references the principles of the BM25 algorithm from the field of information retrieval. By introducing a term frequency saturation function and document length normalization on the basis of traditional TF-IDF, it is more suitable for handling the characteristics of legal texts, capable of identifying potential legal professional terms in queries and assigning them higher weights.

\subsubsection{Legal Knowledge Background Construction Mechanism}
\label{sec3.1.2}

Legal knowledge background construction calculates the relevance of legal concepts to queries through multi-dimensional evaluation, with the formula:

\begin{equation}
\begin{split}
R(C, Q) = {} & \alpha \cdot \left[ \sum_{t \in Q \cap C} IDF(t) \cdot \biggl(\frac{f(t, C) \cdot (k_1+1)}{f(t, C) + k_1 \cdot (1-b+b \cdot \frac{|C|}{avgdl})} \biggr) + \delta \right] \\
             & + \beta \cdot R_{kg}(C, Q) + \gamma \cdot R_{case}(C, Q) + \delta \cdot R_{jur}(C, Q)
\end{split}
\end{equation}

where $R_{text}(C, Q)$ calculates text relevance using the BM25+ algorithm, $f(t, C)$ is the frequency of term $t$ in concept $C$, $|C|$ is the document length of concept $C$, $avgdl$ is the average concept document length, $k_1$ is the term frequency saturation parameter, $b$ is the document length normalization parameter, and $\delta$ is the long document compensation factor. $R_{kg}(C, Q)$ calculates association based on the shortest path length between concept nodes in the knowledge graph, $R_{case}(C, Q)$ calculates case law weight based on concept citation statistics in relevant precedents, and $R_{jur}(C, Q)$ measures jurisdictional relevance through the Jaccard similarity coefficient. Weights $\alpha$, $\beta$, $\gamma$, $\delta$ are determined through Bayesian optimization methods based on historical query data, satisfying the weight normalization condition, implementing multi-dimensional relevance assessment of legal concepts, and providing relevant legal knowledge background for the model.

\subsubsection{Legal Reasoning Guidance and Professional Path Templates}
\label{sec3.1.3}

The legal reasoning guidance framework adopts multi-dimensional assessment of the professional level of responses, including five key dimensions: legal accuracy, content comprehensiveness, citation standardization, logical rigor, and professional expression standardization. Each dimension is assigned different weights, with the final quality score calculated through weighted computation, ensuring generated legal analyses meet professional standards. The assessment standards reference the accuracy dimension in the QUEST assessment framework, focusing on evaluating the consistency of AI system output content with legal authoritative standards.

Legal accuracy assesses concept accuracy scores; content comprehensiveness measures the coverage of required legal points in the response; citation standardization evaluates the format correctness and reliability of legal citations; logical rigor is assessed through coherence scores between adjacent reasoning steps; professional expression standardization evaluates terminology accuracy, format compliance, and style appropriateness. Weights for each dimension are determined by expert Delphi method and satisfy the weight normalization condition, forming a systematic legal response quality assessment framework to guide large language models in generating legal analyses that meet professional standards.

\subsubsection{Dynamic Prompt Optimization Mechanism}
\label{sec3.1.4}

The dynamic prompt optimization mechanism, as an adaptive component of this framework addressing the complexity of legal dispute analysis, improves legal analysis quality through continuous monitoring and feedback adjustment. This mechanism adopts a closed-loop design concept, transforming assessment feedback into prompt optimization instructions, thereby achieving the system's adaptive learning capability. During operation, this mechanism exhibits phased characteristics, with initial generation, quality assessment, dynamic adjustment, and regeneration as core links.

\begin{figure}[htbp]
\centering
\includegraphics[width=120mm]{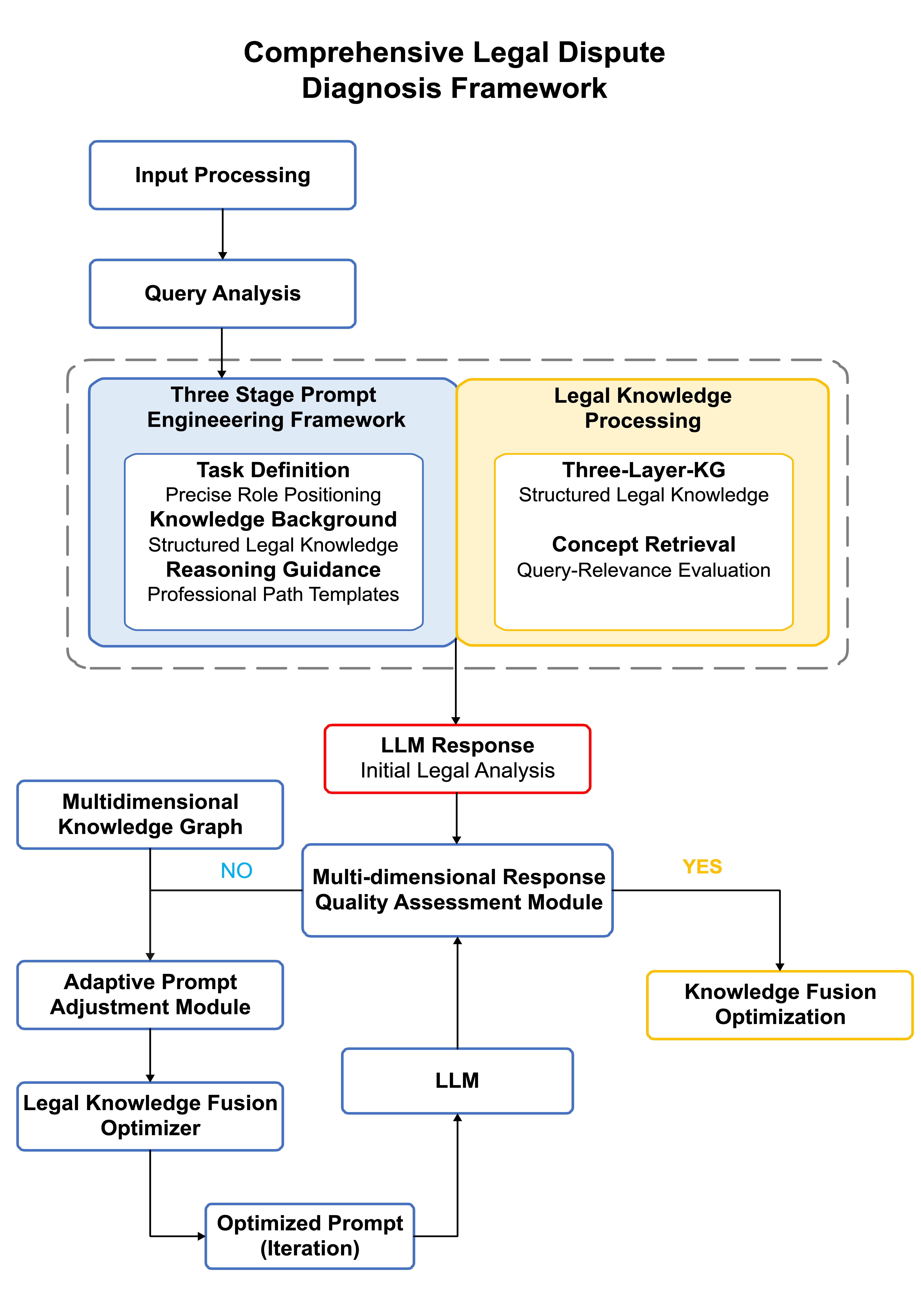}
\caption{Dynamic Prompt Optimization Mechanism Workflow}
\label{fig2}
\end{figure}

Figure~\ref{fig2} illustrates the workflow of the dynamic prompt optimization mechanism. The process begins with the user's legal query input, where the system automatically generates an initial legal prompt based on the query content, guiding the large language model to produce preliminary legal analysis. Subsequently, the multi-dimensional response quality assessment module evaluates the generated content and makes quality judgments based on preset thresholds: if the quality meets requirements, the final legal analysis result is output directly; if the quality does not meet standards, an iterative optimization cycle is initiated. In the optimization cycle, the adaptive prompt adjustment module adjusts the prompt structure and content based on quality assessment feedback, while the legal knowledge fusion optimizer provides professional knowledge in combination with the multidimensional knowledge graph, jointly forming an optimized prompt. The entire process constitutes a closed-loop path of "assessment-adjustment-fusion-reassessment" until the generated result meets preset quality standards. This mechanism is particularly applicable to handling complex cases involving intertwined legal concepts, identifying legal reasoning deficiencies in model responses and improving them through prompt optimization, achieving an evolution from "static prompting" to "dynamic dialogue."

\FloatBarrier
\subsection{Design and Implementation of Multidimensional Knowledge Graph}
\label{sec3.2}

The complexity of legal domain knowledge representation requires multi-level structural design to cover legal concepts and relationships at different abstraction levels~\cite{Liu2023}. The multidimensional knowledge graph designed in this research includes both traditional knowledge graph node and relationship representations and multi-granularity concept retrieval mechanisms, capable of dynamically locating relevant legal concepts based on queries. Legal knowledge graphs constructed based on knowledge-enhanced large language models can significantly improve the accuracy and completeness of knowledge representation, providing a reliable knowledge foundation for legal dispute analysis.

\subsubsection{Three-Layer Architecture Legal Knowledge Graph}
\label{sec3.2.1}

The basic architecture of the legal three-stage prompt engineering framework includes legal classification ontology layer, legal representation layer, and legal instance layer, achieving comprehensive coverage from abstract legal concepts to specific case applications~\cite{Chen2023}. This three-layer architecture design conforms to the hierarchical organizational characteristics of legal knowledge, differing from traditional flat knowledge graphs, and can accurately express hierarchical relationships between legal concepts. Structured legal prompt frameworks can effectively guide models in systematic legal analysis, especially in identifying legal meanings. The importance of multi-level knowledge representation for legal reasoning has been validated in previous research~\cite{Zheng2021}, providing a theoretical foundation for the three-layer architecture design of this research.

\subsubsection{Multi-Granularity Legal Concept Retrieval Mechanism}
\label{sec3.2.2}

The multi-granularity legal concept retrieval mechanism designed in this research integrates four independent and complementary matching strategies, achieving precise dynamic retrieval of legal concepts. These four strategies independently calculate similarity scores, ultimately forming a comprehensive score through weight fusion, thereby improving the accuracy and professionalism of retrieval.

First, the direct matching method based on legal norm coding calculation formula is:

\begin{equation}
CM(C, Q) = \gamma \cdot ExactMatch(code(C), code(Q)) + (1-\gamma) \cdot PartialMatch(code(C), code(Q))
\end{equation}

where $code(C)$ and $code(Q)$ represent the standardized legal codes (such as law article numbers) of concept $C$ and query $Q$ respectively, the $ExactMatch$ function returns a boolean value for complete matching (1 for match, 0 for no match), the $PartialMatch$ function calculates the degree of partial matching (range [0,1]), and $\gamma$ is a weight coefficient balancing the importance of complete and partial matching (0$\leq$$\gamma$$\leq$1).

Second, the similarity calculation formula based on legal domain specialized semantic vectors is:

\begin{equation}
VS(C, Q) = \frac{\vec{v}_C \cdot \vec{v}_Q}{|\vec{v}_C| \cdot |\vec{v}_Q|}
\end{equation}

where $\vec{v}_C$ and $\vec{v}_Q$ are semantic vector representations of concept $C$ and query $Q$ generated through legal domain pre-trained models. This method can capture deeper semantic associations in legal texts, transcending the limitations of surface text matching.

Third, the path reasoning score calculation formula based on legal ontology relationship graph is:

\begin{equation}
PI(C, Q) = \max_{q \in concepts(Q)} \left[ \lambda^{d(C,q)} \cdot \sum_{r \in Path(C,q)} weight(r) \right]
\end{equation}

where $concepts(Q)$ is the set of legal concepts identified from query $Q$, $d(C,q)$ is the shortest path length from concept $C$ to concept $q$ in the legal ontology graph, $\lambda$ is the decay factor of path length (0<$\lambda$<1), $Path(C,q)$ is the shortest path from $C$ to $q$, and $weight(r)$ is the weight of relation $r$ on the path, reflecting the importance of different relation types.

Fourth, the term matching score calculation formula based on professional lexical segmentation is:

\begin{equation}
TM(C, Q) = \frac{\sum_{t \in terms(C)} w_t \cdot Match(t, Q) \cdot ILT(t)}{\sum_{t \in terms(C)} w_t}
\end{equation}

where $terms(C)$ is the set of professional legal terms extracted from concept $C$, $w_t$ is the importance weight of term $t$, $Match(t, Q)$ is the matching degree of term $t$ with query $Q$, with the calculation formula:

\begin{equation}
Match(t, Q) = \alpha_1 \cdot ExactMatch(t, Q) + \alpha_2 \cdot StemMatch(t, Q) + \alpha_3 \cdot SemSim(t, Q)
\end{equation}

where $ExactMatch(t, Q)$ is the exact matching score, $StemMatch(t, Q)$ is the stem matching score, $SemSim(t, Q)$ is the semantic similarity score, and $\alpha_1$, $\alpha_2$, and $\alpha_3$ are weight coefficients satisfying $\alpha_1 + \alpha_2 + \alpha_3 = 1$.

The legal professional weight $ILT(t)$ of the term is calculated as:

\begin{equation}
ILT(t) = \log\left(\frac{freq(t, legal) + \sigma}{freq(t, general) + \sigma}\right) \cdot JurScope(t)
\end{equation}

where $freq(t, legal)$ is the frequency of term $t$ in the legal corpus, $freq(t, general)$ is the frequency in the general corpus, $\sigma$ is a smoothing factor, and $JurScope(t)$ is the judicial application scope coefficient of term $t$.

Ultimately, this research adopts a weighted fusion method to integrate the scoring results of four independent strategies, forming a comprehensive score for legal concepts. The fusion process considers the relative importance of each matching strategy in different legal application scenarios, optimizing weight allocation through machine learning methods, ensuring the sum of weights is 1. This multi-strategy fusion mechanism utilizes the structured characteristics and semantic associations of legal texts, capable of precisely identifying legal concepts relevant to queries, providing a reliable knowledge foundation for legal dispute analysis. The system ranks retrieved legal concepts based on comprehensive score results, prioritizing the most relevant content for users, thereby improving the accuracy and professionalism of legal analysis.

\subsubsection{Integration of Knowledge Graph and Web Search}
\label{sec3.2.3}

The timeliness issue of legal knowledge is a critical technical challenge facing legal large language models, as traditional static knowledge representation struggles to adapt to frequent updates of laws and regulations and the dynamic evolution of judicial interpretations~\cite{Li2024}. This framework constructs a multi-source heterogeneous legal knowledge integration mechanism, implementing the coupling of multidimensional knowledge graphs with professional legal web searches. Legal knowledge graphs, as an intermediate layer between users and large language models, can enhance the legal correctness and citation standardization of model answers.

This mechanism first establishes a unified retrieval interface covering authoritative legal data sources, ensuring retrieved legal information meets timeliness requirements through three mechanisms: jurisdictional identification, legal concept timeliness marking, and change tracking; second, it develops a legal authority assessment model, calculating authority scores based on legal source type, publishing institution hierarchy, and citation frequency, prioritizing high-authority information; finally, it implements semantic fusion of knowledge graph and dynamic retrieval results, forming a legal knowledge service system that both maintains the advantages of structured representation and has real-time update capabilities.

\FloatBarrier
\subsection{System Integration and Collaborative Working Mechanism}
\label{sec3.3}

The legal dispute analysis framework constructed in this research implements the integration of multidimensional knowledge graphs with prompt engineering, forming a closed-loop feedback system~\cite{Sun2023}. From a system architecture perspective, the workflow shown in Figure~\ref{fig1} embodies a complete information processing chain: first, the legal query input by the user undergoes preliminary analysis, and the system identifies key legal issues therein; subsequently, the system activates two parallel processing paths—the prompt engineering path and the knowledge processing path. The prompt path is responsible for constructing structured prompts including task definition and reasoning guidance levels, while the knowledge path provides professional legal knowledge support through a three-layer knowledge graph, concept retrieval, and quality assessment components. The processing results of these two paths converge at the fusion stage, jointly constituting enhanced legal prompts that guide the large language model to generate preliminary legal analysis. The system also designs a quality feedback mechanism, triggering an iterative optimization cycle including prompt optimization and knowledge fusion dimensions when analysis quality does not meet preset standards, ultimately outputting professional legal analysis results. This closed-loop design of "input-analysis-enhancement-generation-assessment-optimization-output" significantly improves the overall efficiency of legal dispute analysis, achieving collaborative enhancement effects of knowledge retrieval technology and prompt engineering, providing robust technical support for intelligent analysis of complex legal issues~\cite{Cui2023}.

\FloatBarrier
\section{Experiments and Evaluation of Legal Dispute Analysis}
\label{sec4}

\subsection{Experimental Setup}
\label{sec4.1}

To evaluate the framework proposed in this research, the COLIEE legal question answering dataset and classic legal cases from important courts in multiple countries were selected as evaluation benchmarks. These precedents cover multiple legal domains including intellectual property, contract law, and tort law, from which 100 pairs of question-answer pairs related to legal disputes were randomly selected as the test set~\cite{Chalkidis2022}. The experiment employed two representative large language models, DeepSeek-R1-Distill-Qwen-32B and Qwen/QwQ-32B, with three control configurations designed:

\begin{enumerate}
\item Baseline Configuration (Baseline): Original models without any legal knowledge enhancement mechanisms;
\item Traditional Configuration (Traditional): Applying basic legal knowledge graphs and traditional single-dimension retrieval mechanisms;
\item Complete Configuration (Complete): The complete framework including three-stage prompt engineering and multidimensional knowledge graphs.
\end{enumerate}

The LexGLUE English Legal Language Understanding Benchmark provides standardized evaluation methods, and the cross-jurisdictional legal judgment prediction benchmark enhances the diversity of evaluation~\cite{Niklaus2021}, jointly constituting the methodological foundation for the experimental evaluation of this research.

All experiments were conducted in computing environments with identical configurations, using NVIDIA A100 (40GB) GPUs, to ensure the fairness and comparability of experimental results.

\subsection{Experimental Results and Analysis}
\label{sec4.2}

\subsubsection{BLEU and ROUGE Metric Evaluation}
\label{sec4.2.1}

This experiment uses BLEU and ROUGE metrics to evaluate the quality of texts generated under different configurations. BLEU (Bilingual Evaluation Underresearch) evaluates the n-gram matching degree between generated text and reference text based on precision, including BLEU-1 (word level), BLEU-2 (word pair level), and BLEU-L (longest common subsequence). ROUGE (Recall-Oriented Underresearch for Gisting Evaluation) evaluates the matching degree between generated text and reference text based on recall, including ROUGE-1, ROUGE-2, and ROUGE-L metrics. The experiment compares the generated answers for each set of legal questions with standard answers provided by legal experts, calculating average scores for each metric.

\begin{table}[htbp]
\centering
\caption{Model Performance Comparison on BLEU and ROUGE Metrics}
\label{tab1}
\renewcommand{\arraystretch}{1.2}
\begin{tabularx}{\textwidth}{X *{6}{c}}
\toprule
\multirow{2}{*}{Metric} & \multicolumn{3}{c}{DeepSeek-R1-Distill-Qwen-32B} & \multicolumn{3}{c}{Qwen/QwQ-32B} \\
\cmidrule(lr){2-4} \cmidrule(lr){5-7}
 & Baseline & Traditional & Complete & Baseline & Traditional & Complete \\
\midrule
BLEU-1 & 0.2409 & 0.1124 & 0.2607 & 0.2306 & 0.1730 & 0.3998 \\
BLEU-2 & 0.1351 & 0.0660 & 0.1474 & 0.1204 & 0.1025 & 0.2432 \\
BLEU-L & 0.1715 & 0.1678 & 0.1796 & 0.1893 & 0.1390 & 0.2129 \\
ROUGE-1 & 0.2779 & 0.1713 & 0.2781 & 0.2705 & 0.2631 & 0.3592 \\
ROUGE-2 & 0.0779 & 0.0657 & 0.0862 & 0.0823 & 0.0835 & 0.1369 \\
ROUGE-L & 0.2306 & 0.2497 & 0.2699 & 0.2432 & 0.2398 & 0.3254 \\
\bottomrule
\end{tabularx}
\end{table}

Analysis of the data in Table~\ref{tab1} indicates that traditional configuration scores on BLEU and ROUGE metrics are generally lower than baseline configuration, such as DeepSeek-R1-Distill-Qwen-32B model's BLEU-1 decreasing from baseline's 0.2409 to 0.1124 (a 53.3\% decrease), and ROUGE-1 from 0.2779 to 0.1713 (a 38.4\% decrease). This phenomenon verifies the limitations of single prompt methods—they emphasize format standardization and citation precision but limit the model's ability to generate multi-angle, multi-level legal reasoning. In contrast, complete configuration outperforms baseline configuration on all metrics, such as Qwen/QwQ-32B model's BLEU-1 increasing from baseline's 0.2306 to 0.3998 (a 73.4\% increase), and ROUGE-2 from 0.0823 to 0.1369 (a 66.3\% increase), indicating that the three-stage prompt structure provides the model with sufficient thinking space and analytical perspectives, enabling it to achieve deeper legal reasoning while maintaining professional format, resulting in a higher matching degree between generated text and legal expert standard answers.

\FloatBarrier
\subsubsection{Key Legal Performance Indicator Evaluation}
\label{sec4.2.2}

This experiment uses three indicators—sensitivity, specificity, and precision—to evaluate model performance on legal judgment tasks. Sensitivity measures the model's ability to correctly identify legally relevant issues, calculated as true positives/(true positives + false negatives); specificity measures the model's ability to correctly exclude irrelevant issues, calculated as true negatives/(true negatives + false positives); precision measures the proportion of actually relevant issues among issues judged by the model to be relevant, calculated as true positives/(true positives + false positives). The test samples in this research were annotated by three senior legal experts as the standard for evaluation.

\begin{table}[htbp]
\centering
\caption{Evaluation of Legal Large Language Models on Key Performance Indicators}
\label{tab2}
\renewcommand{\arraystretch}{1.2}
\begin{tabularx}{\textwidth}{X *{6}{c}}
\toprule
\multirow{2}{*}{Indicator} & \multicolumn{3}{c}{DeepSeek-R1-Distill-Qwen-32B} & \multicolumn{3}{c}{Qwen/QwQ-32B} \\
\cmidrule(lr){2-4} \cmidrule(lr){5-7}
 & Baseline & Traditional & Complete & Baseline & Traditional & Complete \\
\midrule
Sensitivity & 0.6201 & 0.5867 & 0.6892 & 0.6546 & 0.6319 & 0.7285 \\
Specificity & 0.7841 & 0.7532 & 0.8266 & 0.8062 & 0.7834 & 0.8547 \\
Precision & 0.6041 & 0.5638 & 0.6566 & 0.6262 & 0.6018 & 0.7047 \\
\bottomrule
\end{tabularx}
\end{table}

The experimental results in Table~\ref{tab2} reveal performance differences among different configurations in legal judgment tasks. Traditional configuration scores on all three key indicators are lower than baseline configuration, such as DeepSeek-R1-Distill-Qwen-32B model's sensitivity decreasing from 0.6201 to 0.5867 (a 5.4\% decrease), specificity from 0.7841 to 0.7532 (a 3.9\% decrease), and precision from 0.6041 to 0.5638 (a 6.7\% decrease). This performance decline confirms the structural limitations of single prompts, which emphasize direct citation of legal provisions and precedents while neglecting in-depth consideration of complex relationships between legal concepts and consideration of different legal perspectives. This tendency limits the model's judgment ability when facing cases with unclear legal boundaries and complex disputes. In contrast, complete configuration achieves significant improvements on all indicators, such as Qwen/QwQ-32B model's sensitivity increasing from baseline's 0.6546 to 0.7285 (an 11.3\% increase), indicating that the three-stage prompt structure and legal reasoning path templates provide the model with an analysis framework combining structured guidance with flexibility, enhancing its judgment ability in complex legal contexts.

\FloatBarrier
\subsubsection{Legal Content Evaluation Dimension Analysis}
\label{sec4.2.3}

This experiment evaluates the quality of legal content generated by models from three dimensions: citation accuracy, reasoning rationality, and conclusion reliability. Citation accuracy measures the accuracy of model citations of legal provisions and precedents; reasoning rationality evaluates the logic and rationality of model legal reasoning; conclusion reliability measures the reliability of model conclusions. The evaluation adopts a 0-1 scoring standard, with scores averaged after independent scoring by five legal experts, ensuring the objectivity and accuracy of the evaluation.

\begin{table}[htbp]
\centering
\caption{Performance of Legal Large Language Models on Legal Content Evaluation Dimensions}
\label{tab3}
\renewcommand{\arraystretch}{1.2}
\begin{tabularx}{\textwidth}{X *{6}{c}}
\toprule
\multirow{2}{*}{Dimension} & \multicolumn{3}{c}{DeepSeek-R1-Distill-Qwen-32B} & \multicolumn{3}{c}{Qwen/QwQ-32B} \\
\cmidrule(lr){2-4} \cmidrule(lr){5-7}
 & Baseline & Traditional & Complete & Baseline & Traditional & Complete \\
\midrule
Citation Accuracy & 0.58 & 0.71 & 0.81 & 0.61 & 0.69 & 0.79 \\
Reasoning Rationality & 0.64 & 0.59 & 0.77 & 0.67 & 0.61 & 0.81 \\
Conclusion Reliability & 0.66 & 0.57 & 0.84 & 0.70 & 0.62 & 0.86 \\
\bottomrule
\end{tabularx}
\end{table}

The legal content quality evaluation results presented in Table~\ref{tab3} reveal an important observation: traditional configuration shows improvement in citation accuracy, such as DeepSeek-R1-Distill-Qwen-32B model increasing from baseline's 0.58 to 0.71 (a 22.4\% increase), but decreases in reasoning rationality and conclusion reliability, with reasoning rationality decreasing from 0.64 to 0.59 (a 7.8\% decrease) and conclusion reliability from 0.66 to 0.57 (a 13.6\% decrease). This "strong citation, weak reasoning" performance distribution validates this research's hypothesis: single prompts emphasize format standardization and citation completeness, causing models to tend toward "knowledge stacking" rather than "knowledge fusion," while lacking clear guidance for legal reasoning paths, hindering models from establishing complete legal analysis frameworks. In contrast, complete configuration achieves improvements in all three dimensions, such as Qwen/QwQ-32B model's citation accuracy increasing from baseline's 0.61 to 0.79 (a 29.5\% increase), indicating that the three-stage prompt structure achieves a transition from "simple application" to "comprehensive integration," integrating legal knowledge acquisition and reasoning path guidance, enabling the model to conduct legal analysis and conclusion inference while maintaining citation precision, providing more comprehensive intelligent assistance for legal dispute analysis.

\FloatBarrier
\subsubsection{Case Research Analysis: Liebeck v. McDonald's Hot Coffee Burn Case}
\label{sec4.2.4}

\begin{figure}[htbp]
\centering
\includegraphics[width=\textwidth]{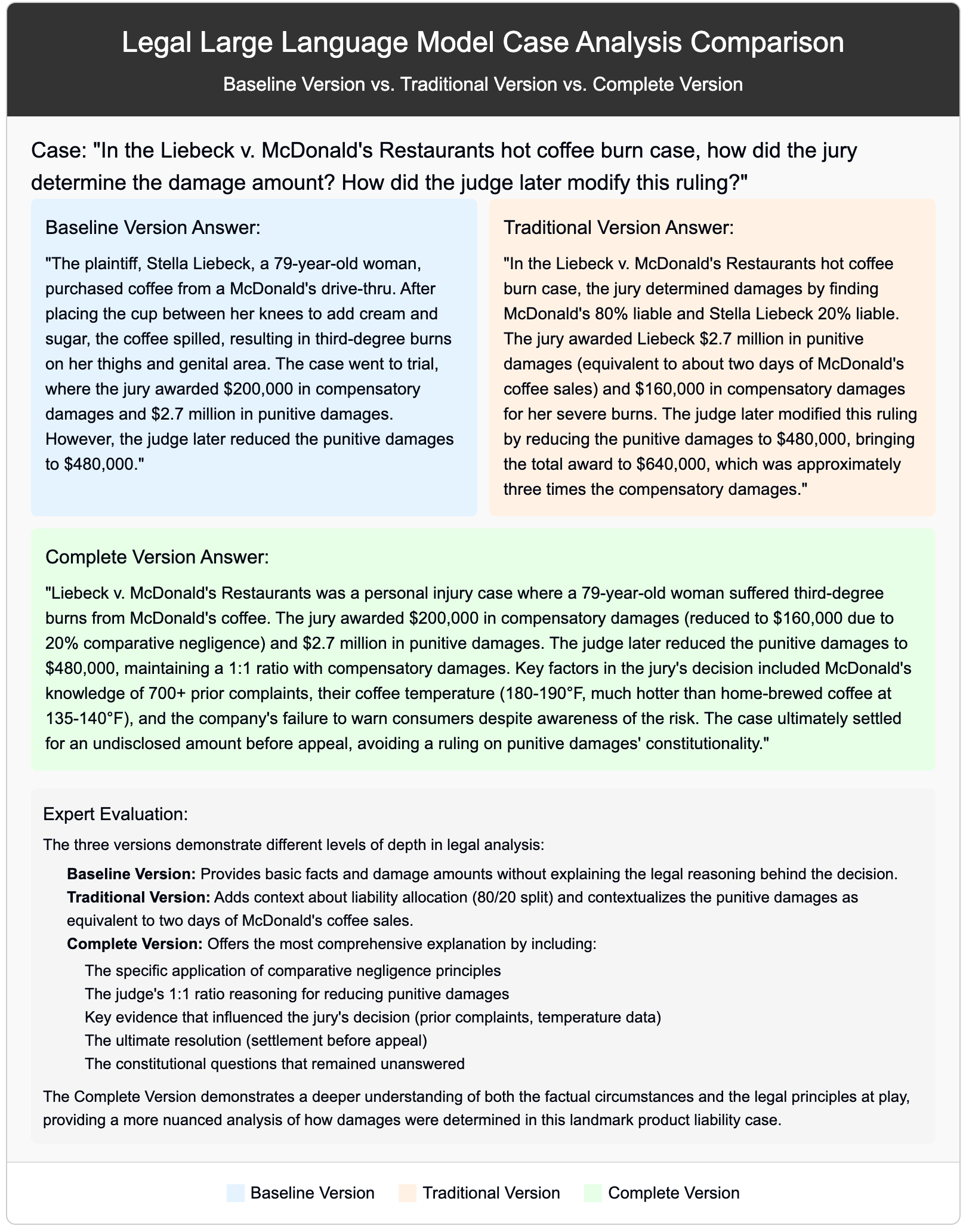}
\caption{Legal Large Language Model Case Analysis Comparison: Baseline Version vs. Traditional Version vs. Complete Version}
\label{fig3}
\end{figure}

To verify the effectiveness of the framework proposed in this research in handling actual legal cases, the famous Liebeck v. McDonald's hot coffee burn case was selected as a typical test case. This case involves multiple complex legal issues such as liability determination, compensation calculation, and judicial discretion, making it an appropriate sample for evaluating the depth of legal analysis. The three configurations demonstrated different analytical capabilities when handling this case, with their differences presented in Figure~\ref{fig3}.

The comparative analysis shown in Figure~\ref{fig3} demonstrates the capability differences among the three configurations when handling the Liebeck v. McDonald's hot coffee burn case. The baseline configuration's response is limited to basic facts and simple data presentation, providing compensation amount information (\$200,000 compensatory damages and \$2.7 million punitive damages, later reduced by the judge to \$480,000), but lacking elaboration on legal principles and reasoning processes. The traditional configuration builds on the baseline by adding liability distribution ratios (McDonald's 80\% responsibility, Liebeck 20\% responsibility) and contextualizing the punitive damages amount (equivalent to McDonald's two-day coffee sales), but still does not deeply analyze the legal logic chain. In contrast, the complete configuration provides comprehensive legal analysis, covering the specific application mechanisms of comparative negligence principles, legal basis for the 1:1 compensation ratio principle, inferential value of key evidence (over 700 complaint records and coffee temperature data), procedural evolution of case resolution methods, and related constitutional controversy issues. From a legal analysis structural perspective, the complete configuration's response closely resembles the classic IRAC (Issue, Rule, Analysis, Conclusion) analysis framework in Anglo-American legal memoranda; from a content depth perspective, it reveals the critical role of customary evidence in the jury's "reflective balancing" decision-making process, and the case's evolution trajectory from "first-order results" (specific rulings) to "second-order results" (case law influence). Legal expert evaluation confirms that the complete configuration not only provides more accurate legal application analysis but also demonstrates a nuanced grasp of judicial decision-making logic, validating the effectiveness and adaptability of this research's three-stage prompt structure and multidimensional knowledge graph fusion framework in handling complex legal cases.

\FloatBarrier
\subsection{Manual Evaluation Results of Legal Dispute Analysis}
\label{sec4.3}

In the manual evaluation stage, this research adopted the QUEST framework evaluation indicators, including information quality, understanding and reasoning, expression style and role, safety and harm, and trust and confidence, with each dimension using a 1-5 scoring standard. The evaluation was independently completed by 10 legal domain experts (including law professors, senior judges, and practicing lawyers), providing a comprehensive evaluation of 30 typical legal dispute cases, with the average score for each dimension as the final score.

\begin{table}[htbp]
\centering
\caption{Model Scores on Legal QUEST Framework Evaluation Indicators (1-5 points)}
\label{tab4}
\renewcommand{\arraystretch}{1.2}
\begin{tabularx}{\textwidth}{X *{6}{c}}
\toprule
\multirow{2}{*}{Indicator} & \multicolumn{3}{c}{DeepSeek-R1-Distill-Qwen-32B} & \multicolumn{3}{c}{Qwen/QwQ-32B} \\
\cmidrule(lr){2-4} \cmidrule(lr){5-7}
 & Baseline & Traditional & Complete & Baseline & Traditional & Complete \\
\midrule
Information Quality & 3.6 & 3.2 & 4.1 & 4.1 & 3.7 & 4.4 \\
Understanding \& Reasoning & 3.3 & 2.9 & 3.8 & 4.0 & 3.6 & 4.4 \\
Expression Style \& Role & 3.8 & 3.4 & 4.2 & 4.2 & 3.9 & 4.5 \\
Safety \& Harm & 4.0 & 3.6 & 4.5 & 4.3 & 4.0 & 4.6 \\
Trust \& Confidence & 3.5 & 3.0 & 4.0 & 4.0 & 3.5 & 4.5 \\
\bottomrule
\end{tabularx}
\end{table}

The manual evaluation results presented in Table~\ref{tab4} further confirm the limitations of single prompt methods in legal dispute analysis tasks. Traditional configuration performs lower on all QUEST evaluation dimensions, such as DeepSeek-R1-Distill-Qwen-32B model scoring 2.9 in the "Understanding \& Reasoning" dimension (12.1\% lower than baseline) and 3.0 in the "Trust \& Confidence" dimension (14.3\% lower than baseline), indicating that single prompts limit the depth and reliability of model legal reasoning, resulting in reduced trust from legal experts in system outputs. Legal expert evaluation points out that content generated by traditional configuration, while containing legal terminology and citations, presents a "superficially professional, substantively shallow" characteristic when solving actual legal problems, with the limitations of single prompts emphasizing knowledge acquisition while neglecting reasoning guidance becoming more apparent in complex legal dispute cases. In contrast, complete configuration achieves significant improvements in all evaluation dimensions, demonstrating the advantages of the three-stage prompt engineering and multidimensional knowledge graph fusion framework, achieving a transition from "legal knowledge application" to "legal wisdom expression," with generated content that is not only professionally accurate but also possesses sophisticated legal reasoning and perspective consideration, enhancing legal experts' recognition of artificial intelligence systems in legal dispute analysis.

\section{Conclusion}
\label{sec5}

This research proposes an enhanced framework for legal dispute analysis that integrates prompt engineering with multidimensional knowledge graphs to address fundamental limitations in legal language models. The framework establishes a three-stage hierarchical prompt structure (comprising task definition, knowledge background, and reasoning guidance components) that works synergistically with a three-layer knowledge graph architecture to form a closed-loop system for legal analysis. Experimental validation confirms the framework's effectiveness across multiple dimensions, with the complete configuration demonstrating improvements in both automated metrics and expert evaluations. The framework achieves enhanced text generation quality, superior legal judgment capabilities (with sensitivity increasing by 7.3-11.3\%, specificity by 5.4-6.0\%, and precision by 8.7-12.5\%), as well as higher quality legal content in terms of citation accuracy, reasoning rationality, and conclusion reliability.

Future research will expand this framework in three primary directions to accommodate broader legal application scenarios. First, cross-language legal dispute analysis capabilities will be developed to address multinational legal applications and jurisdictional differences. Second, integration of multimodal legal evidence analysis will enhance the system's ability to process diverse documentation formats, including visual and audio evidence. Finally, improving legal reasoning explainability will increase the transparency and trustworthiness of the system's analytical processes, potentially addressing artificial intelligence accountability concerns in legal contexts.

\section*{Acknowledgments}
This work was supported by the Special Fund for the Central Government to Guide Local Science (Grant No. 202407AB110003), the Key Research and Development Program of Yunnan Province (Grant No. 202402AA310056), the National Natural Science Foundation of China (Grant No. 62366057), and the Scientific Research Fund Project of the Yunnan Provincial Department of Education (Grant No. 2024J0024).


\end{document}